# Detecting Violence in Video Based on Deep Features Fusion Technique

Heyam M. Bin Jahlan[1][0000-0001-6141-2445] and Lamiaa A. Elrefaei[1,2][0000-0001-5781-2251]

[1] Computer Science Department, Faculty of Computing and Information Technology, King Abdulaziz University, Jeddah 21589, Saudi Arabia

[2] Electrical Engineering Department, Faculty of Engineering at Shoubra, Benha University, Cairo 11629, Egypt

hjahlan@stu.kau.edu.sa, laelrefaei@kau.edu.sa, lamia.al-refaai@feng.bu.edu.eg

**Abstract.** With the rapid growth of surveillance cameras in many public places to monitor human activities such as in malls, streets, schools and, prisons, there is a strong demand for such systems to detect violence events automatically. Automatic analysis of video to detect violence is significant for law enforcement. Moreover, it helps to avoid any social, economic and environmental damages. Mostly, all systems today require manual human supervisors to detect violence scenes in the video which is inefficient and inaccurate. in this work, we interest in physical violence that involved two persons or more. This work proposed a novel method to detect violence using a fusion technique of two significantly different convolutional neural networks (CNNs) which are AlexNet and SqueezeNet networks. Each network followed by separate Convolution Long Short Term memory (ConvLSTM) to extract robust and richer features from a video in the final hidden state. Then, making a fusion of these two obtained states and fed to the max-pooling layer. Finally, features were classified using a series of fully connected layers and softmax classifier. The performance of the proposed method is evaluated using three standard benchmark datasets in terms of detection accuracy: Hockey Fight dataset, Movie dataset and Violent Flow dataset. The results show an accuracy of 97%, 100%, and 96% respectively. A comparison of the results with the state of the art techniques revealed the promising capability of the proposed method in recognizing violent videos.

## 1  Introduction

The deployment of camera surveillance in most popular places, attract computer vision researchers to develop systems to detect human activities. The recent development in the field of human action recognition has led to a renewed interest in the detection of specific action which is violence. Automatic violence detection is basically involved in increasing the safety of humans by monitoring their behavior, in addition to prevent social, economic and environmental damages. Until now, most of camera surveillance systems are supervised by the human to manually analyze visual information and detect violence behavior. This manual supervision is practically infeasible and inefficient which leads to strong demand for automated violence detection systems. For this purpose, various studies have been proposed to automatically recognize human activities as in [1][2] and suspicious activities in surveillance videos as in [3][4].

Violence acts can be defined as a specific abnormal event that involved one or more persons with physical force to damage something or hurt a human [5]. Based on the literature, the basic processes to detect violence in the video are: first, divide the whole video into frames to extract frame-level features, then, the location of persons in frames was specified in order to detect motions of persons over time [5]. After that, spatial features are extracted either by learning techniques or hand-crafted techniques. Finally, classified extracted features using classification techniques [6].

Features are a crucial component to detect activities in videos and it is discriminative information extracted from the image. Recently, many techniques have emerged in order to address the problem of action recognition. Most of these techniques based on deep features and hand-crafted features [5]. The handcrafted feature is designed for a specific problem and is developed by the researcher, for this reason, this approach is expensive and unsatisfied in real-world problems.

Recently, the rapid growth of various deep learning techniques has been helped in improving the field of computer vision in addition to the availability of huge data and computational resources. Most techniques that addressing some problems in computer vision like action recognition, object detection, tracking, etc., have been gained high performance using deep learning techniques. However, much of the recent literature has not used deep learning-based techniques to address the problem of violence detection from video. Previous studies have only focused on hand-crafted techniques to represent visual information of videos, unfortunately, these methods fail to achieve a better result when handling unseen data, alternatively, deep learning-based techniques have the ability to achieve a high degree of generalization compared to handcrafted features. Also, deep learning is not designed for a specific problem and it can be applied to different tasks using the same architecture. Moreover, the input to deep learning can be raw pixels and don't require complex preprocessing [7].

## 1.1   Action Recognition

Action recognition represents a big umbrella that includes recognition of particular action which is violence. In action recognition purpose, the hand-crafted feature can be holistic representation features as [8][9] or local representation features [10] [11]. holistic representation is based on a global representation of the body's movement and structure [12]. In [8] authors suggest to extract key-frames from the video using the probabilities computation of each frame and select one with high probability. Another method in [9], authors propose a new descriptor called Spatio-temporal Laplacian pyramid coding (STLPC) to extract features from the whole video. Whereas local representation focused to extract spatial and temporal interest points in the frame for instance, in [10] authors propose a new approach motion key point trajectory (MKT) and used novel descriptor called trajectory-based covariance (TBC) in order to tracking motion points. Authors in [11] propose a novel motion descriptor called trajectory based on low level local feature.

On the other side, deep neural networks such as Convolutional Neural Network (CNN) have become the better choice in learning the contents of images. Authors in [13] propose to use a 3D Convolution Neural Network 3D (CNN) rather than 2D to extract spatial and temporal features from video to capture the motion information. Their architecture consists of multiple channels from a set of frames and performs convolution and subsampling to each channel respectively to get a final feature by combining all channels. In [14] authors represent Spatio-temporal information of skeleton sequences as color texture images, called joint distance maps (JDMs), then using the Convolution Neural Network (CNN) to extract the invariance feature from it. In [15] authors show a novel method called three-channel deep convolutional neural networks (3ConvNets) + weighted hierarchical depth motion maps (WHDMM) that consist from sequences of depth maps as the input to the ConvNets. In [16] authors suggest a Robust Non-Linear Knowledge Transfer Model (R-NKTM) is a deep fully connected neural network transfers knowledge of human actions from an unseen view to a shared high-level view through finding a set of non-linear transformations that connects the views.

## 1.2 Violence Detection

many hand-crafted techniques have been proposed to detect violence. we can restrict the features used to detect violence into some popular features such as optical flow, space-time interest points and motion binary pattern [17]. One common type of optical flow descriptor that suggested to detect violence in the crowd scenes is proposed in [18] which named Violent Flows (ViF) that calculates the statistics of how the flow vectors magnitude changes over the time to detect the change of violence and non-violence behavior. This ViF then classified using a linear SVM classifier. The extension of this work [19] has been suggested Oriented Violent Flows (OViF) to calculate the orientation of flow vector in addition to the magnitude values. Another method in [20] that aimed to detect and localized violence in video using the Gaussian Model of Optical Flow (GMOF) then, Spatio-temporal features are extracted from each candidate location using Orientation Histogram of Optical Flow (OHOF). The study in [21] suggests a new descriptor of Improved Fisher Vectors (IFV) to represent the video. Their method suggests extracting Spatio-temporal features using Improved Dense Trajectories (IDT). Then each descriptor is represented using IFV and finally classified by linear SVM.

Another feature is space-time interest point (STIP) and defined as analyzing the variation of space and time of interest points in sequences of frames [17]. In [22] authors suggested a new method called three-layered Bag-of-Visual-Words (BoVW). This framework is consisting of three layers first is low-level, second is mid-level and third is high level, the low layer is aimed to describe the contents of the video, the mid-layer is to coding and pooling the output of low-level whereas the high level includes the classification task using Support Vector Machine (SVM). In [23] authors introduce a bag of words framework using Space-Time Interest Points (STIP) and Scale-Invariant Feature Transform (MoSIFT) features.

The third and common feature used to detect violence in the video is the Motion Binary Pattern (MBP) which assumes that the change of intensity of particular pixel over time indicates the motion occurred in that pixel. In [24] authors suggest Region Motion Vector (RMV) descriptor by first extract motion vector from a compressed video sequence. Then, analyzed the attributes of these motion vector in each frame and between frames. This RMV classified using a radial basis SVM. The specialized Lagrangian technique has been suggested in [25] which is used the information of background motion compensation, appearances and long-term motion. In [26] authors used extreme acceleration features using 2D Fast Fourier Transform descriptor on consecutive frames of video. This can be done by computing the direction and magnitude of acceleration. Finally, these obtained features are classified using SVM and AdaBoost classifiers. In [27] authors suggest a new method to extract features by detecting the blob of movement then describing these blobs assuming that the violence blobs have a specific shape and position.

Hand-crafted based approaches achieve good accuracy but on the other hand, deep learning can directly act and automatically extract features. The method in [7] used CNN to extract spatial features from the frame. Then, each feature map extracted is accumulated using Convolution Long Short Term Memory. The input to CNN is a difference between two adjacent frames. Features then classified using a fully connected and softmax classifier. In [5] authors propose a novel method by first, detecting the object using the CNN model. Then, passing frames with individual detected to 3D CNN, where the spatiotemporal features are extracted. Finally, features are classified using a softmax classifier. In [28] authors present bi-channels convolutional neural network (CNN) method one designed for the original frame to extract appearance features and another channel for the differential of adjacent frames to extract motion features. Each channel then applied to different linear SVMs to classify the feature. Finally using the label fusion technique. Authors in [29] propose Spatiotemporal Encoder built on the Bidirectional Convolutional LSTM (BiConvLSTM) architecture. They extract feature maps from each frame using CNN then, pass these feature maps to a BiConvLSTM to extract temporal information by both a pass forward in time and in reverse. Next, applied an elementwise maximization to represent the video. Finally, pass this representation to a classifier to identify whether the video contains violence.

Other studies used a hybrid approach of deep learning and handcrafted approaches, for instance, in [30] authors propose a novel method to extract features from the video using Features from Accelerated Segment Test (FAST) detector and Binary Robust Invariant Scalable Key Points (BRISK) descriptor to summarized the video in the single representative image. Then, the Hough forest was used as a classifier. In [31] authors utilize advantages of both handcrafted and deep learning features by suggesting two convNet models based on deep learning to extract spatial-temporal features and trajectories in videos to capture more information.

Most of previous studies have been focused on hand-crafted approach which is expensive and inefficient in real-world problem. On the other hand, most of deep learning approaches aimed to extract Spatio-temporal features. However, researches do not take into account using a fusion of significantly different CNN model to extract more robust features.

**Our contributions can be summarized as follows:**

- This work proposed a novel method to detect fighting violence using fusion of deep features technique.

- Extract two feature vectors from significant different CNN models. First, a simplified version of AlexNet that implemented with less number of parameters and without Local Response Normalization layers [32] and second is SqueezeNet model [33].

- Each feature vector is applied to separate Convolution Long Short Term Memory to aggregate features into the final hidden state.
- Using a fusion between these final states to capture richer and robust features.

The remainder of this paper is organized as follows: the proposed method is introduced in Section 2; then, experimental results are presented in Section 3. The paper is concluded in Section 4.

## 2 Proposed Framework

In this section, we intend to provide a detailed description of the proposed method which is aimed to develop end to end model to classify the video to violence or non-violence. first, the camera will capture the video stream, then, we applied some preprocessing to this video before applying to CNN such as select key frames using threshold technique and resize each frame to 224x224x3. After that, spatial features were extracted from each frame using two different channels, each channel has a separate CNN network followed by convolution long short term memory. We used AlexNet [32] in the first channel beside, SqueezeNet [33] in another channel. Finally, classify the fusion of two final states in each channel using a series of fully connected layers and a softmax layer.

### 2.1 Preprocessing

The input to our model is a short video clip. Before classifying this video into violence or non-violence, we preprocess this short video. First, we select key frames by extracting a sequence of S frames from each video. The whole frames were extracted from the video then, subtract every two adjacent frames and compared with a specific threshold value T. if the count of non-zero values in the difference image is greater than T, then take the current frame assuming that it is significantly different from its previous frame. After that, we resized the image to 224 by extracting a region of size 224 by 224 pixel's square from the input image randomly.

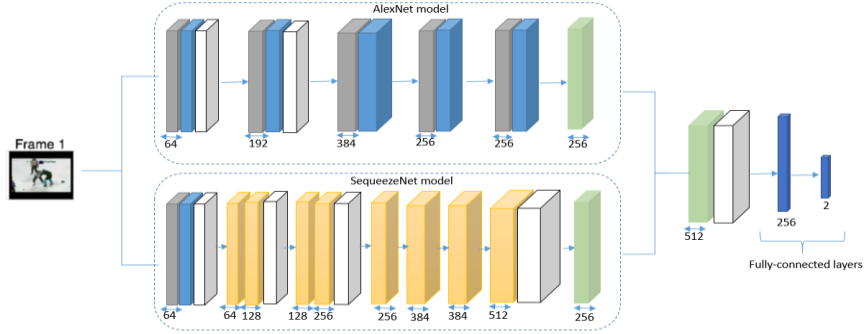

**Fig. 1.** Block diagram of the proposed model. AlexNet model consists of alternating convolution (gray), normalization (blue) and pooling (white) layers. SqueezeNet below consists of alternating convolutional (gray), normalization (blue), fire (yellow) and pooling (white) layers. The hidden state of the ConvLSTM (green). The numbers under each convolution, fire and hidden state layers refer to dimension of feature map. The fully-connected layers are shown in dark blue color.

### 2.2 Feature extraction with Convolution Neural Networks models

For feature extraction purpose, we use two significantly different pre-trained CNN networks, AlexNet [32] and SqueezeNet [33]. The input to these networks is a sequence of S frames. The architecture of each network will be described in details in the following subsections:

**AlexNet Architecture.** Due to a lot of computation when using original AlexNet model [34], this lead to use a simplest implementation of AlexNet that have some of properties [35] [36]:

1. Reduced number of parameters to avoid a lot of computation.
2. This model doesn't use Local Response Normalization (LRN) as in the original paper.
3. This model uses Adaptive average pooling layer to process different size of images.

In the first layer of this network, the input for AlexNet is a 224x224x3 RGB image which passes through the first convolutional layer with 64 filters having size 11×11 and a stride of 4. The image dimensions' changes to 55x55x64. Then the AlexNet applies a maximum pooling layer with a filter size 3×3 and a stride of two. The resulting image dimensions will be reduced to 27x27x64. In the second convolutional layer with 192 feature maps having size 5×5 and a stride of 1. Then, there is again a maximum pooling layer with filter size 3×3 and a stride of 2. the output will be reduced to 13x13x192.

The third, fourth and fifth layers are convolutional layers with filter size 3×3 and a stride of one. The third layer used 384 feature maps whereas the third and fourth used 256 filters. The three convolutional layers are followed by a maximum pooling layer with filter size 3×3, a stride of 2 and have 256 feature maps. In this work, we removed the last fully connected layers in order to extract convolutional features. Figure 1 illustrates the architecture of this network in the proposed method.

**SqueezeNet Architecture.** Smaller CNN architecture achieves AlexNet-level accuracy on ImageNet with 50x fewer parameters. There are three main strategies when designing this CNN architecture:

- Replace 3x3 filters with 1x1 filters.
- Decrease the number of input channels to 3x3 filters
- Down-sampling late in the network so that convolution layers have large activation maps to increase accuracy.

SqueezeNet begins with a standalone convolution layer (conv1), followed by 8 Fire modules (fire2-9), ending with a final conv layer (conv10) for classification. gradually increase the number of filters per fire module from the beginning to the end of the network. SqueezeNet performs max-pooling with a stride of 2 after conv1, fire4, fire8, and conv10 layers; these relatively late placements of pooling are per on third strategy.

A Fire module is comprised of: a squeeze convolution layer (which has only 1x1 filters), feeding into an expand layer that has a mix of 1x1 and 3x3 convolution filters [33]. In this work, we removed the final conv10 layer in order to replace it with a fully connected layer and softmax classifier. Figure 1 illustrates the architecture of this network in the proposed method.

## 2.3   Feature Aggregation with Convolutional Long Short Term Memory

Each channel consists of CNN followed by Convolution Long Short Term Memory. CNN network is well-suitable to extract spatial features of frames. The video consists from a sequence of frames. To detect the temporal information in the video, we need to detect the location of persons and the motion of them across time. For this purpose, The Recurrent Neural Network (RNN) is a good choice. We used a special variation of RNN called a convolutional Long Short Term Memory (ConvLSTM) that proposed in [37]. This ConvLSTM has replaced the fully connected gates in ordinary LSTM with convolutional gates. These convolutional gates have a capable of encoding temporal changes by aggregating features of the current frame with its neighbor frame. We used 256 filters with size 3x3 and stride1. Figure 2 illustrates the inner architecture of ConvLSTM. The equations of the ConvLSTM model are given in equations 1-5. Where ($*$) refers to the convolutional operator, ($\circ$) refers to Hadamard product, ($\sigma$) sigmoid layer, ($W$) represents the weight, ($X$) is the input feature, ($b$) represents the bias. The gate activations $i_t$, $f_t$ and $o_t$, the memory cell $C_t$ and the hidden state $H_t$ are all 3D tensors [37].

$$i_t = \sigma(W_{xi} * X_t + W_{hi} * H_{t-1} + W_{ci} \circ C_{t-1} + b_i) \quad (1)$$
$$f_t = \sigma(W_{xf} * X_t + W_{hf} * H_{t-1} + W_{cf} \circ C_{t-1} + b_f) \quad (2)$$
$$C_t = f_t \circ C_{t-1} + i_t \circ \tanh(W_{xc} * X_t + W_{hc} * H_{t-1} + b_c) \quad (3)$$
$$o_t = \sigma(W_{xo} * X_t + W_{ho} * H_{t-1} + W_{co} \circ C_t + b_o) \quad (4)$$
$$H_t = o_t \circ \tanh(C_t) \quad (5)$$

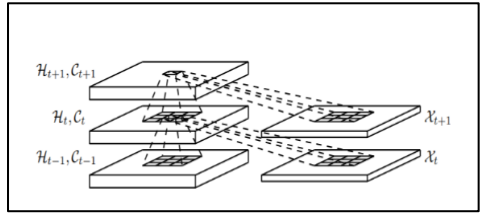

**Fig. 2.** Inner architecture of ConvLSTM [37].

### 2.4 Classification

The output from each channel is the final hidden state of dimension 256x6x6 that aggregate whole frame' features. The next step is to make a fusion of these two hidden states. The dimension of the resulting state is 512x6x6. This resulting state then fed to a max-pooling layer which results in the dimension of 512x3x3. This state is flattened to turns them into a single vector then classified into violence or non-violence using a series of fully-connected layers. We have only two fully connected layers that take the input of features and return the probabilities for each label.

## 3 Experiments and Results

Our model is implemented using the Pytorch library. We perform various experiments on a gaming device with NVIDIA GeForce 1050TI. The operating system was Windows 10 using 6 core 2.2 GHZ i7-8750 H and 16 GB RAM.

### 3.1 Datasets

We evaluated our proposed model on three benchmark public datasets, Hockey Fight [23], Movie [23], and Violent Flow [18] datasets. Table 1 illustrates a detailed description of each dataset.

**Table 1.** detailed description of public datasets.

| Dataset | Number of samples | Resolution | Description |
|---|---|---|---|
| Hockey Fight [23] | 1000 videos, 500 fight and 500 non-fight | 360x288 | Hockey dataset contains short videos of the National Hockey League (NHL) game. the main challenges are: blurring, camera movement and there are other people not involved to violence in the video. |
| Movie Dataset [23] | 200 videos, 100 fight and 100 non-fight | 360x250 | Fight videos are collected from action movies and non-fight are collected from available public datasets for action recognition. The main challenge is the significantly different in its content and resolution. |
| Violent Flow Dataset [18] | 246 videos, 123 fight and 123 non-fight | 320x240 | Videos are collected from YouTube. Most of videos are collected from football matches. The main challenge is videos have a small part of crowed that involved violence while a large part remained as spectators. |

### 3.2 Data Augmentation

The data augmentation plays an important role in order to avoid overfitting our model by increasing the number of samples in the dataset. All deep learning models achieve a better result with large datasets. In our work, we use some data transformation on training set only such as applied horizontal flip by 50/50 chance, rotate the image by 20 angles, and normalize frame by making a mean equal to zero and variance unity. In the validation and testing sets, we only normalize frames. Figure 3 show the sample of each transformation technique.

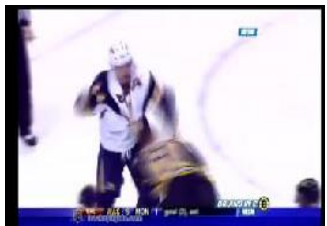 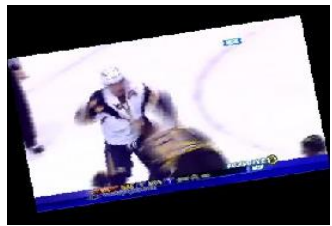

(a)  (c)

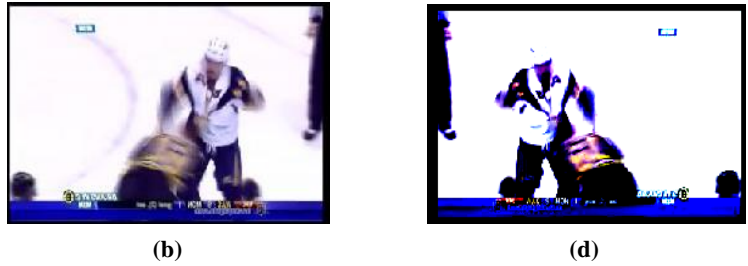

**Fig. 3.** (a) original frame. (b) horizontal flip (c) rotate frame 20 angel (d) normalization

### 3.3 Experimental Result and Discussion

The whole samples in the dataset has been divided randomly into three sets: training set, validation set, and testing sets. For this purpose, we select 60% of data for the training set and 20% for the validation set and 20% for the testing set. In the following sections, we will demonstrate various experiments that applied to the training set and the results obtained in the testing set.

**Training stage.** To training our model we experiment with different parameters such as batch size, learning rate, weight decay, sequence length, number of epochs and, an optimization algorithm to achieve better performance. The best accuracy was obtained using 40 epochs for Hockey and 20 epochs for Movie dataset, whereas in the Violent Flow dataset, we use 60 epochs to achieve good performance, learning rate with 1e-4 and RMSprop optimization algorithm. A batch size of 8 video clips was used as an input with 20 sequence length. The weight decay was set to 0.05. Cross entropy loss was used during training. Feature extracted using 20 frames that applied to AlexNet and SqueezeNet networks. The output obtained from AlexNet is a feature map of dimension 256x3x3. After ConvLSTM layer, the output becomes 256x6x6. Whereas the feature map obtained from SqueezeNet is 512x13x13. Before passing this feature map to ConvLSTM layer, we applied it to a max-pooling layer to put the width and height as same as the feature map obtained from AlexNet channel which is equal to 6x6. So, the input to ConvLSTM layer is a feature map of dimension 512x6x6, and an output of dimension 256x6x6. After extract these two feature maps of dimension 256x6x6, we concatenate them into one feature of dimension 512x6x6. Finally, to reduce the dimension, we add a max-pooling layer to get a feature map of 512x3x3. In order to evaluate our model in real world, we used a hold-out validation schema by evaluating our model in the same validation set after each 5 epochs and save the model that achieves high accuracy in the validation set. Figure 4 shows the accuracy and loss values of the training and validation sets in each dataset.

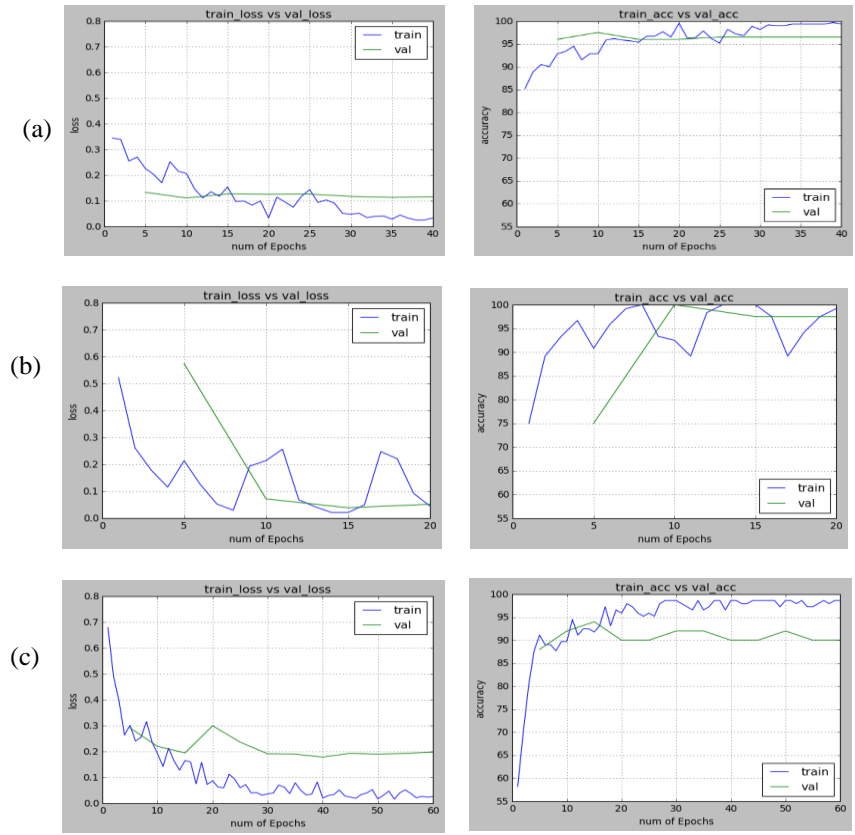

**Fig. 4.** First column represents loss values; second column represents accuracy values. (a) Hockey dataset, (b) Movie dataset, (c) violent Flow dataset

**Testing Stage.** To test our model, we used hold-out validation by evaluating model after each 5 epochs using same validation set and saving model that achieves high accuracy in the validation set to evaluate the model in the testing set. The accuracies achieved were 97%, 100%, and 96% in Hockey, Movie, and Violent Flow datasets respectively. The confusion matrix of each dataset is presented in figure 5.

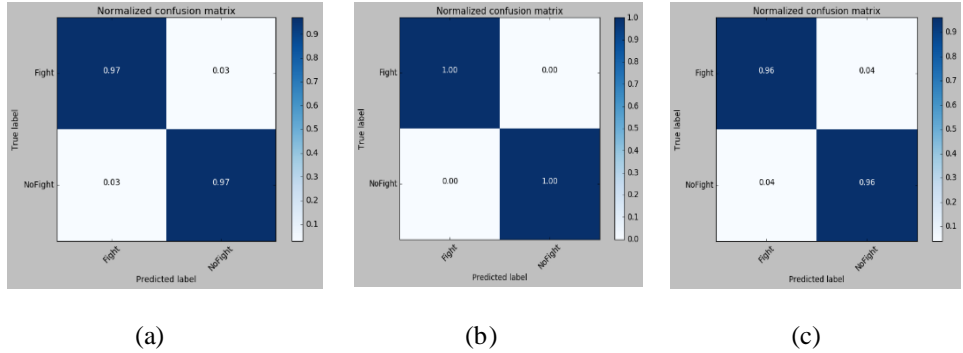

(a) (b) (c)

**Fig. 5.** Confusion matrix of (a) Hockey Fight (b) Movie (c) Violent Flow datasets

### 3.4 Comparative Discussion

The comparative of the proposed method with state of the art methods based on the accuracy of detection is presented in this section. Table 2 shows the results of some recent studies in addition to the proposed method. In [28] authors suggest bi-channel for extract appearance and motion features and use SVM to classify these features. They obtained the accuracy of 95% and 93% in Hockey and Violent Flow datasets respectively. Another method [19] proposed to use the magnitude and orientation of ViF descriptor and classified using SVM. They achieved accuracies 87.5% and 88% for Hockey and Violent Flow datasets respectively. In [29] the feature extracted using CNN and BiConvLSTM. They obtained 96.96%, 100% and 90% in Hockey, Movie and Violent Flow dataset respectively. In method [30] features extracted from the video using Features from Accelerated Segment Test (FAST) detector and Binary Robust Invariant Scalable Key Points (BRISK) descriptor Then, the Hough forest was used as a classifier. The accuracies achieve were 94.6% and 99% in Hockey and Movie datasets respectively. In [31] authors utilize two convNet models based on deep learning to extract spatial-temporal features and trajectories in videos then classified using SVM. They achieved accuracies of 98% and 92% in Hockey and Violent Flow datasets respectively. A comparison of the results with the state of the art techniques revealed the promising capability of the proposed method over most of previous techniques.

**Table 2.** compare our method with state of the art method

| Ref. | Feature extraction | Classification | Datasets Accuracy(%) | | |
|---|---|---|---|---|---|
| | | | Hockey | Movie | Violent Flow |
| Xia et al. [28] (2018) | VGG-f model | SVM | 95% | -- | 93% |
| Gao et al. [19] (2016) | ViF+OViF | SVM | 87.5% | -- | 88% |
| Hanson et al. [29] (2019) | CNN+ Bi-ConvLSTM | softmax | 96.96% | 100% | 90.6% |
| Serrano et al. [30] (2018) | Three streams + LSTM | softmax | 94.6% | 99% | -- |
| Meng et al. [31] (2017) | Trajectories + Convolution feature | SVM | 98% | -- | 92% |
| Proposed Method | Two different CNN networks+ ConvLSTM | softmax | 97% | 100% | 96% |

## 4 Conclusion

In this paper, we introduce the learning-based model to detect violence in videos. Our proposed method is summarized into three steps First, in preprocessing, we use a threshold technique to extract key frames from the video. Secondly, features extracted from the frame using two significantly different CNN models AlexNet and SqueezeNet. These features then aggregated using Convolution Long Short-Term Memory to obtain temporal information. Furthermore, we make a fusion between two states in order to get deeper and robust features. In the last stage for classification features into violence and non-violence, we use a series of fully connected layers and softmax classifier. We conducted our experiment on three benchmark datasets Hockey Fight, Movie and Violent Flow datasets and achieve better performance over state of the art methods. We obtained accuracies with 97%,100% and 96% in Hockey, Movie, and Violent Flow datasets respectively.